\documentclass[10pt,journal,compsoc]{IEEEtran}

\usepackage{hyperref}       
\usepackage{url}            
\usepackage{booktabs}       
\usepackage{amsfonts}       
\usepackage{nicefrac}       
\usepackage{microtype}      
\usepackage{graphicx}

\usepackage{amsmath,amsthm,amssymb}
\usepackage{verbatim}
\usepackage{subcaption}
\ifCLASSOPTIONcompsoc
  \usepackage[nocompress]{cite}
\else
  \usepackage{cite}
\fi

\newtheorem{theorem}{Theorem}
\newtheorem{proposition}{Proposition}
\newtheorem{lemma}{Lemma}
\newtheorem{definition}{Definition}

\ifCLASSOPTIONcompsoc
  \usepackage[nocompress]{cite}
\else
  \usepackage{cite}
\fi

\begin{document}

\title{Learning on Hypergraphs with Sparsity}

\author{Canh~Hao~Nguyen~\IEEEmembership{}
        and~Hiroshi~Mamitsuka~\IEEEmembership{Member,IEEE}
\IEEEcompsocitemizethanks{\IEEEcompsocthanksitem Canh Hao Nguyen and Hiroshi Mamitsuka are with Institute for Chemical Research, Kyoto University\protect\\
E-mail: \{canhhao,mami\}@kuicr.kyoto-u.ac.jp
}}

\IEEEtitleabstractindextext{%
\begin{abstract}
Hypergraph is a general way of representing high-order relations on a set of objects. It is a generalization of graph, in which only pairwise relations can be represented. It finds applications in various domains where relationships of more than two objects are observed. On a hypergraph, as a generalization of graph, one wishes to learn a smooth function with respect to its topology. A fundamental issue is to find suitable smoothness measures of functions on the nodes of a graph/hypergraph.  We show a general framework that generalizes previously proposed smoothness measures and also gives rise to new ones. To address the problem of irrelevant or noisy data, we wish to incorporate sparse learning framework into learning on hypergraphs. We propose \emph{sparsely smooth formulations} that learn smooth functions and induce sparsity on hypergraphs at both hyperedge and node levels. We show their properties and sparse support recovery results. We conduct experiments to show that our sparsely smooth models have benefits to irrelevant and noisy data, and usually give similar or improved performances compared to dense models.
\end{abstract}

}

\maketitle
\IEEEdisplaynontitleabstractindextext

\IEEEpeerreviewmaketitle

\IEEEraisesectionheading{\section{Introduction}\label{sec:introduction}}

Hypergraphs, representing higher-order relationships,  can be found in broad applications: multi-label learning \cite{Sun08}, computer vision \cite{OchsB12}, biology \cite{Weighill15}, information retrieval \cite{Bu10}, social networks \cite{Cook06} and scientific literature \cite{Newman10}. Graphs and hypergraphs are very useful in data analysis, by which currently are of much interest in machine learning,  particularly when dealing with high-dimensional data and complicated distributions \cite{Chapelle10,Ma11}.  In learning on hypergraphs, following the same vein as on graphs,  smooth functions play the key role \cite{Chung97,Zhu03} in a variety of problem settings  including, but not limited to: smooth label functions in supervised learning \cite{Kyng15}, smooth cluster assignment \cite{Zhou06,Buhler09} and  regularization term in semi-supervised learning \cite{Zhu03}, in which labels are expected to be smooth on high density area.

On graphs, smoothness measures are mainly based on the graph Laplacian \cite{Chung97} and its variants \cite{Buhler09,Kyng15}. However, generalizing to smoothness measures on hypergraphs poses new challenges. So far, in many methods, smoothness measures on hypergraphs have been studied simply by reducing hypergraphs to graphs using clique or star expansions  \cite{Zien99,Agarwal06}. However, since these expansions are not one-to-one mappings, the original hypergraph topology is lost, potentially resulting in information loss and other undesirable effects such as graphs with different node degrees. Another method is total variation  \cite{Hein13}, which is later shown to be inflexible for several reasons. Therefore, it is desirable to have new smoothness measures for hypergraphs with different desirable properties.

One of the main problems in machine learning is that data can be noisy or irrelevant to the problem. This could happen with hypergraphs and it is desirable to handle it explicitly.  For example, in predicting house price, data can come in form of hypergraph, of which each hyperedge is a group of houses. While a hyperedge containing houses in the same neighborhood is \emph{relevant} to the problem in the sense that their prices are expected to be similar (i.e. prices are \emph{smooth} on the hyperedge), a hypergraph for houses whose owners are of the same ethnic group might be \emph{irrelevant}. In recommending music for users on social networks, while communities (as hyperedges) with the same listening pattern might be relevant, communities with the same workplace might not. This is inherently similar to irrelevant and noisy features problem in multivariate data.  In such data, with additive models, sparse learning \cite{Tibshirani96,Friedman07} is a solution due to its ability to make the models sparse, potentially removing irrelevant and noisy data automatically. Lasso \cite{Tibshirani96} would be able to eliminate irrelevant features to the problem and its variants \cite{Yuan06,Jacob09,Zhou10} can detect noisy features within a given group using appropriate sparse-inducing regularization terms.  In a similar manner, to eliminate irrelevant hyperedge and noisy nodes, we wish to incorporate sparsity into hypergraphs that automatically detect \emph{irrelevant hyperedges} and \emph{noisy nodes}. In hypergraph setting, sparsity is meant to be that  (\textbf{i}) the label is very smooth on only a small number of hyperedges (highly non-smooth on others), or (\textbf{ii}) the label is very similar on some nodes, different on others within a single hyperedge. These concepts are to be defined formally later on.

In this work, we first present \emph{a unified  framework} that generalizes all other smoothness measures for graphs/hypergraphs, such as graph Laplacian-based learning, star and clique expansions \cite{Agarwal06} or total variation \cite{Hein13}. The framework allows to analyze if a measure can be achieved with  expansions, or whether it is useful for keeping hypergraph information in its formulation. The framework is general enough to be used to design new smoothness measures.  We then derive sparsely smooth formulations, which learn functions on the nodes of a hypergraph that are \emph{smooth on only a subset} of nodes or hyperedges in a hypergraph.  We further present theoretical properties of  learning on hypergraphs that statistical consistency of sparse support recovery (\emph{sparsistency}) of these models can be achieved. Finally, we conducted experiments to show that our formulations allow us with great flexibility in supervised learning on hypergraphs.  Overall, the contributions of this work are: 
\begin{enumerate}
\item We present a unified framework for the problem of learning smooth functions on hypergraphs that covers all previously proposed smoothness measures.
\item We propose new models for learning smooth functions with sparsity on hypergraphs at both hyperedge and node levels. 
\item We show the theoretical properties of our formulations, including computational efficiency and  consistency of sparse support recovery. We also empirically examine the performance of several models in this framework and show the benefit of sparse models.
\end{enumerate}

This paper is organized as follows. Section 2 describes the unified
smoothness measure framework, which is a generalization of
previous work. We show a key property that  learning on hypergraph cannot be
achieved by simply using graphs. Section 3 and 4 devotes to formulation and theoretical analyses of learning on hypergraph
with sparsity. We then show simulations to show properties of these 
formulations. Sections 6 shows experiments on real data sets to
show how sparsity can help improve supervised learning performances. We
then conclude the paper.

\section{Smoothness Measures on Hypergraphs}

\begin{figure}
 \begin{center}
   \includegraphics[width=0.3\textwidth]{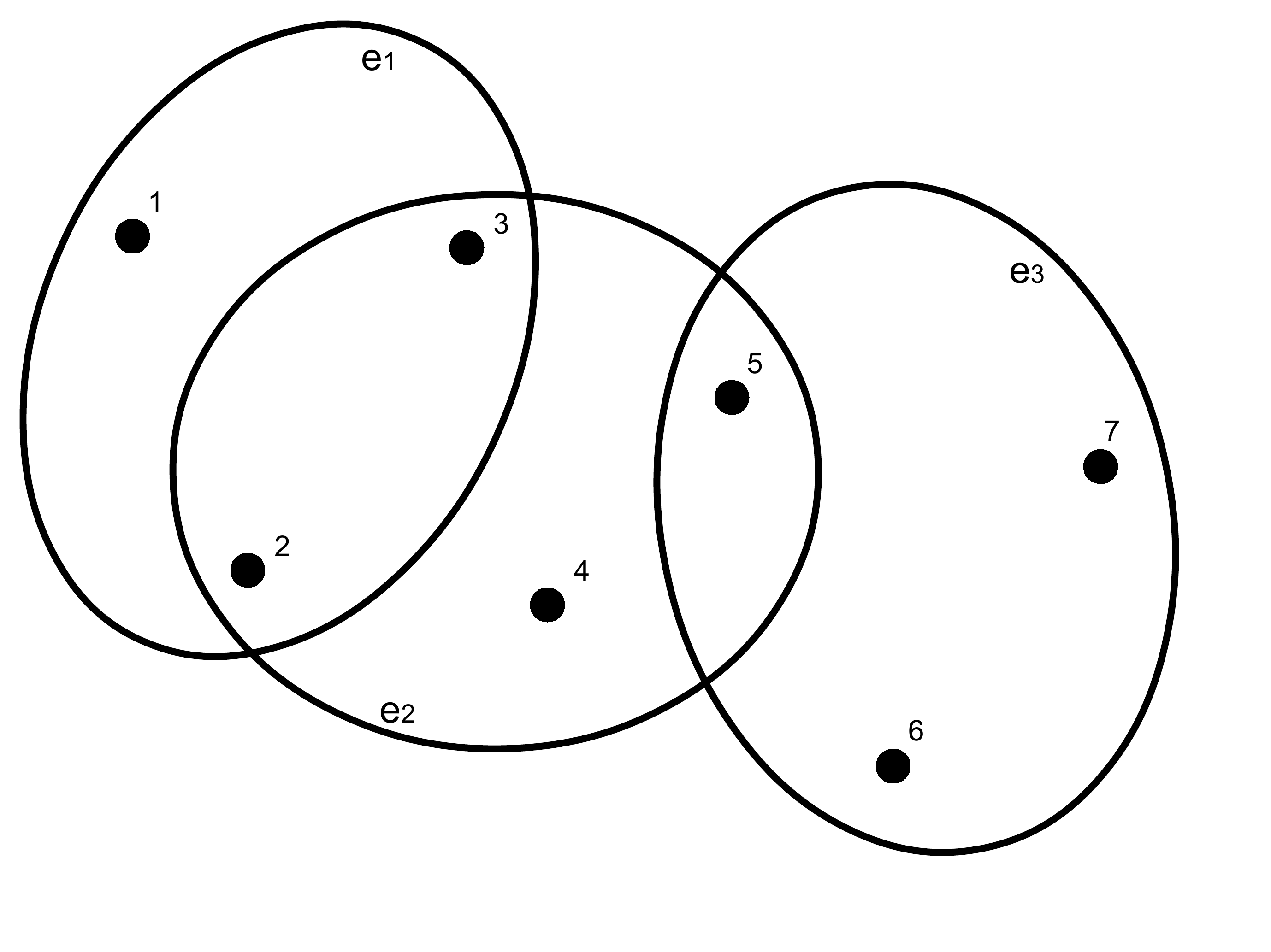}  
   \end{center}
   \caption{Hypergraph example with nodes numbered from 1 to 7. Hyperedges are the node subsets  $e1$, $e2$ and $e3$.} 
\label{fig:exp1}
\end{figure}

A hypergraph $G$ is defined as $G = (V,E)$ comprising of a node set $V$ and hyperedge set $E$ as depicted in Figure \ref{fig:exp1}. Each hyperedge is a subset of the nodes,  $E = \{e_k\}^m_{k=1}$ and $E \subseteq 2^V$. Cardinality $|V| = n$ and $|E| = m$. Each hyperedge $e \in E$ is a subset of the nodes with size  $|e|$ and weight $w(e)$ (could be given or set appropriately for each method). Note that a graph is a special hypergraph with the special requirement that each hyperedge is of size 2. It is commonly assumed that the set of hyperedges encode similarity relationship among the nodes in $V$. 

Similar to learning on graphs, for the similarity relationship in hypergraphs, it is expected that  the nodes in the same hyperedge tend to have similar labels, such as the same class, the same cluster assignment or similar regression values. Therefore, the label function to be learnt will vary less within a hyperedge. These label functions will be called \emph{smooth} on the graph or hypergraph. This is why smoothness measures is at the core of learning problems on graphs and hypergraphs: to score how much a label function varies within the hyperedges. 

\begin{table*}[t]
\caption{Different variants of (\ref{genobj}) as smoothness measures for graphs/hypergraphs. Unreferenced formulations are newly derived from this framework.}
\label{tab:allfor}
\begin{center}
\begin{tabular}{|c|c|c|c|c|}
\hline
type & $T$ & $t$ & $s(f_i,f_j)$ & $sh(f)$   \\      \hline
graph & $\sum$ & . & $(f_i-f_j)^2$ & $ f^TLf$: graph Laplacian \cite{Chung93}   \\
graph & $\sum$ & . & $|f_i-f_j|^p$ & $ <f,\Delta_pf>$: p-Laplacian \cite{Buhler09}  \\
graph & $\sum$ & . & $|f_i-f_j|^{p\to \infty}$    &  Lipschitz extension (lex-minimizer) \cite{Kyng15}  \\
graph & $\max$ & . & $|f_i-f_j|$    & Lipschitz extension (inf-minimizer) \cite{Kyng15} \\
hypergraph & $\sum$  & $\sum$ & $(f_i-f_j)^2$ &  $f^TLf$ of clique/star expansion  \cite{Agarwal06} \\
hypergraph & $\sum$  & $\sum$ & $|f_i-f_j|$ &  clique expansion + 1-Laplacian  \\
hypergraph & $\sum$  & $\max$ & $|f_i-f_j|$ &  total variation \cite{Hein13} \\
hypergraph & $\max$  & $\max$ & $|f_i-f_j|$ &  inf-minimizer + star/clique expansion  \\
hypergraph & $\max$  & $\sum$ & $|f_i-f_j|$ &  max hyperedge smoothness \\
hypergraph & any & $(\sum \cdot)^{\frac{1}{p}}$ & $|f_i-f_j|^p$ & within-hyperedge $l_p$ norm  \\
hypergraph & $(\sum \cdot)^{\frac{1}{p}}$ & any &  $|f_i-f_j|^p$ & between-hyperedge $l_p$ norm  \\
\hline
\end{tabular}
\end{center}
\end{table*}

We show that all previously proposed smoothness measures of a function $f : V \to R$ on graph or hypergraph $G$, with $f_i$ being the value on node $i$,  has the following unified form (as in Table \ref{tab:allfor}):
\begin{equation}\label{genobj}
sh(f) =   T_{e \in E} ( t_{i,j \in e}(w_{ij} \cdot  s(f_i,f_j))),
\end{equation}
\begin{itemize}
\item $sh(f)$:  smoothness measure of  function $f$ (on $G$),
\item  $T(.)$: the function combining  smoothness measures on all hyperedges,
\item  $t(.)$: the function combining  smoothness measures on all pairs of nodes belonging to a hyperedge,
\item $s(f_i,f_j)$: smoothness measure on the pair of nodes $(i,j)$, and
\item $w_{ij}$: weight of pair of nodes $(i,j)$.
\end{itemize}
In other words, smoothness measure of a function on a hypergraph is combination ($T$) of smoothness measures on each hyperedge, which, in turn, is combination ($t$) of smoothness measures on all pairs of nodes in the hyperedge. The combinations, at both hypergraph or hyperedge levels, can take many forms such as  $\sum$, $\max$, $\l_p$ and other possibilities.

There are many variants of smoothness measures by simply choosing different $T$ and $t$. We list them in Table \ref{tab:allfor} and show their corresponding names and interpretations. When $T$ and $t$ are both  $\sum$ or $\max$, the smoothness measures of a hypergraph can be expressed in the smoothness measures of its star or clique expansion graph. All the other cases listed in the Table, $sh$ cannot be computed on expansion graphs. For example, when $T= \sum$, $t = \max$, then $sh(f)$ becomes the sum of maximum differences in all hyperedges, a.k.a total variation. The reason $sh(f)$ cannot be computed in its star or clique expansion graphs is that hyperedge information is no longer available in graphs. This kind of measures is unique to hypergraph setting and useful for keeping hypergraph information (which star and clique expansions would fail to keep). It can be seen that these variants captures all previously proposed smoothness measures to the best of our knowledge. There are also many other variants that have not been proposed before. Therefore, within this general framework, we could propose many new smoothness measures, shown as the ones without references in the table. 

\section{Sparsely Smooth Models}

While there are many  existing and newly proposed smoothness measures for hypergraphs as in the previous Section, it is the main objective  of this work to study the cases of smoothness measures with sparsity property, to show different sparsity patterns and their usefulness. Sparsity is an exciting topic in machine learning and statistics, used to improve interpretability, robustness, effectiveness and efficiency of learning models \cite{Tibshirani13}. It have been used to select features in additive models such as lasso \cite{Tibshirani96} and many extensions in many situations. Similar to learning additive models, we consider the problem that there are many hyperedges that are irrelevant to the problem or  within a hyperedge, many nodes are noisy. This is a practical setting as one may have data of a hypergraph without knowing which hyperedges are useful for the prediction problem in hand. In this case, one wishes to learn a function $f$ on (the node set of) a hypergraph that:
\begin{enumerate}
\item $f$ is only smooth on a subset of the hyperedge set $E_r \subseteq E$, not smooth on the rest $E_i$ ($ = E \backslash E_r$). In this case, we say that $f$ has the property of \emph{hyperedge-wise sparsity} or \emph{hyperedge selection}.
\item $f$ is only smooth on a  subset of the nodes in a hyperedge. We say that $f$ has the property of \emph{node-wise sparsity} or \emph{node selection}.
\end{enumerate}

\textbf{Definition.} Any function $f$ with either of the two properties is called  \textbf{sparsely smooth}. 
Ideally, smoothness measure can be  $0$ on some hyperedges/nodes, so that these hyperedges/nodes are considered relevant to the problem while  the other hyperedges/nodes are not. However, this is not usually possible in practice. The solution is to have  a (nonzero) cut-off threshold that  separates smooth and non-smooth hyperedges or nodes, which we call \emph{hyperedge or node selection ability}, respectively.  

\subsection{Supervised Learning} 
In the scope of this paper, we specifically deal with supervised learning problem. We work on regression setting, with smoothness measures are used as regularization. Classification, semi-supervised learning can be done by simply replacing the label fitting term in the objective function. For regression, we are given a hypergraph $G$ on which we know the label of some node encoded in label vector $L \in \mathbb{R}^n$.  We also assume that in the hypergraphs, some hyperedges are not supposed to be relevant to the learning problem. To learn sparsely smooth functions on a hypergraph, we formulate the following optimization problem
\begin{equation}\label{genobj2}
f^{*} = \arg \min_{f \in R^n} \frac{1}{2} |f-Y|^2_L + \lambda sh(f).
\end{equation}
with $\lambda$ is the tradeoff parameter between faithfulness to the known label  $Y$ and smoothness measures. $|.|_L$ is the difference between the known and predicted labels only \emph{on the nodes with known labels}. $T$ and $t$ are chosen so that $sh$  has hyperedge and/or node selection abilities. Since constraining a fixed number of nodes/hyperedges usually results in computationally inefficient formulations, we use $l_1$ norm in similarly to lasso \cite{Tibshirani96}. We use a lasso-like formulation, which can be regarded as star expansion, equipped with $l_1$ norm for sparsity while preserving convexity.

\subsection{Sparsely Smooth Models} 
We propose new formulations using a slack variable $\mu_k$ for each hyperedge $e_k \in E$, which can be interpreted as star expansion-styled formulations and $\mu_k$ becomes the label of star node of $e_k$, which plays the role of mean/central/representative label of a hyperedge. Note that while the formulations  we propose are still mathematically equivalent to, but not of exactly the same form as the general form of smoothness measure in (\ref{genobj}). We propose to use the following formulation:
\begin{equation}\label{genobj1}
sh(f) =   T_{e_k} ( ss(f,e_k))
\end{equation}
where $ss(f,e_k)$ is the smoothness measure on hyperedge $e_k$. $ss(f,e_k)$ plays the same role as  $t_{i,j \in e}(w_{ij} \cdot  s(f_i,f_j))$ in (\ref{genobj}). We show some examples of $ss(f,e_k)$ that we will use later. 

\textbf{Case 1.}  $ss_1(f,e_k) = \min_{\mu_k \in R} w(e_k)\sum_{i \in e_k} |f_i-\mu_k|$.  $ss_1(f,e_k)$ is the sum of absolute differences between the labels of nodes in hyperedge $e_k$ and a value $\mu_k$ that is chosen to minimize the sum. This smoothness measure shows the total difference of the labels from $\mu_k$, which play the role of a representative label of the hyperedge. 

\textbf{Case 2.} $ss_2(f,e_k) =  \min_{\mu_k \in R} w(e_k)\max_{i \in e_k} |f_i-\mu_k|$. $ss_2(f,e_k)$ is the maximum absolute difference of all the labels of nodes in $e_k$ from a value $\mu_k$ that is chosen to minimize the maximum difference. A property of this measure is that $\mu_k$ becomes the midpoint of the maximum and minimum labels of the nodes. In other words, 
\begin{align*}\min_{\mu_k \in R} ss_2(f,e_k) &= \frac{1}{2} w(e_k) (\max_{i \in e_k}(f_i) - \min_{i \in e_k} (f_i)) \\ &= \frac{1}{2} w(e_k) \max_{i,j \in e_k} |f_i-f_j|.\end{align*}
Note that this is half of total variation \cite{Hein13}.

\begin{definition}  \emph{Hyperedge selection}: our efficient version of total variation \cite{Hein13}, the formulation is 
\begin{align}\label{starhyperedgesel}
f^{*} &= \arg\min_{f \in R^n, \delta_k, \mu_k\in R} \frac{1}{2} |f-Y|_L^2  +  \lambda \sum_{k=1}^m  w(e_k)  \delta_k \\ \nonumber
s.t  &\ \   |\mu_k-f_i| \le \delta_k, \forall i \in e_k, k=1..m.
\end{align}
\end{definition}

At the optimal solution of (\ref{starhyperedgesel}), slack variable $\delta_k = \max_{i \in e_k} |\mu_k-f_i|$  and $\mu_k$ is set to minimize $\delta_k$. This makes $w(e_k)  \delta_k = ss_2(f,e_k)$ as in the aforementioned case 2.  Hence, $\mu_k$ becomes the mid-range of $f_i,\ \forall i \in e_k$. $\delta_k$  becomes half of their range and $\delta_k$ is half of total variation. This formulation is mathematically equivalent to total variation regularization (with appropriate $\lambda$) using star expansion style with $\mu_k$ plays the role of label of star node \cite{Hein13}. This is more efficient than original total variation formulation as analyzed later on. This formulation has hyperedge selection ability in the sense that having $ \sum_{k=1}^m  w(e_k)  \delta_k$, which is similar to $l_1$ norm of the vector $(\delta_k)_{i=1}^m $, many $\delta_k$ would become $0$, and the remaining become large. One can select hyperedges that have very smooth labels on them.


\begin{definition} \emph{Node selection}: the  formulation is 
\begin{align}\label{starnodesel2}
f^{*} &= \arg\min_{f \in R^n, \delta_k, \mu_k\in R} \frac{1}{2} |f-Y|_L^2  +  \lambda \sum_{k=1}^m w(e_k) \delta_k^2 \\ \nonumber
s.t  &\ \  \sum_{i \in e_k} |\mu_k-f_i| \le \delta_k,  k=1..m.
\end{align}
\end{definition}
At the optimal solution, slack variable $$\delta_k = \sum_{i \in e_k} |\mu_k-f_i| = \frac{ss_{1}(f,e_k)}{w(e_k)},$$ 
and $\mu_k$ becomes the median of $\{f_i\}_{i \in e_k}$. Using $l_1$ norm constraint, many of the $|\mu_k - f_i|$ would be close to zero, giving this formulation node selection ability. Having $\delta_k^2$ would not make many $\delta_k = 0$, therefore no hyperedge selection ability.

\begin{definition}  \emph{Joint hyperedge and node selection}: the formulation is 
\begin{align}\label{starjointsel}
f^{*} &= \arg\min_{f \in R^n, \delta_k, \mu_k\in R} \frac{1}{2} |f-Y|_L^2  +  \lambda \sum_{k=1}^m w(e_k) \delta_k \\ \nonumber
s.t  &\ \   \sum_{i \in e_k} |\mu_k-f_i| \le \delta_k,  k=1..m.
\end{align}
\end{definition}
Similar to (\ref{starnodesel2}), at the optimal solution, slack variable $$\delta_k = \sum_{i \in e_k} |\mu_k-f_i| = \frac{ss_{1}(f,e_k)}{w(e_k)}.$$ Having $|\mu_k - f_i|$ would have node selection ability in the sense that many $|\mu_k - f_i|$ would become $0$. Different from (\ref{starnodesel2}), having $\sum_{k=1}^m w(e_k) \delta_k$ in the objective function, this formulation has hyperedge selection ability in a similar manner to (\ref{starhyperedgesel}). Table \ref{tab:allsparse} summarizes sparse-inducing property of formulations.

\begin{table}[ht]
\caption{Comparison of smoothness measures and their sparse-inducing property.}
\label{tab:allsparse}
\begin{center}
\begin{tabular}{|c|c|c|c|c|c|}
\hline
Selection & $T$ & $ss(f,e_k)$ \\ \hline
none &  $\sum_k $ & $\sum_{i \in e_k} |\mu_k-f_i|^2$ \\
hyperedges &  $\sum_k $ & $\max_{i \in e_k} |\mu_k-f_i|$ \\
node  & $\max_k$ & $\sum_{i \in e_k} |\mu_k-f_i|$ \\
joint  & $\sum_k $ & $\sum_{i \in e_k} |\mu_k-f_i|$\\
\hline
\end{tabular}
\end{center}
\end{table}

\subsection{Properties}\label{properties}

\textbf{Computation.} All these formulations are quadratic programs, can be solved efficiently with off-the-shelf softwares. Note that these formulations, similar to star-expansion, incur $m$ more variables to optimize. However, instead of computing all $f_i-f_j$ in a hyperedge, we use  $|\mu_k-f_i|$ constraints, reducing from $\frac{|e_k|(|e_k| -1)}{2}$ to $|e_k|$ terms, with $|e_k|$ being the number of nodes in the hyperedge. For the problem having hyperedges of large cardinalities, star expansion-styled formulations are recommended. 

\textbf{Regularization parameter $\lambda$.} It is noteworthy that too large $\lambda$ will lead to many smooth hyperedges/nodes. The reason is that large $\lambda$ forces smoothness measures of all hyperedges to be small, making the gap between those of relevant hyperedges and of irrelevant hyperedges small. The small gap is then dominated by random error of labels, making relevant and irrelevant hyperedges indistinguishable.  Hence, different from additive models \cite{Tibshirani96}, sparsely smooth models need small enough $\lambda$. This is purely for sparse support recovery purpose. It is noteworthy that too small $\lambda$ is not for generalization purpose that requires $\lambda$ not to be arbitrarily small (or large). Finding the balance is a key issue, but is out of the scope of this paper. 

\textbf{Related formulations.} Our proposed formulations are corresponding to two well-known formalisms in sparse learning:  generalized adaptive fused lasso \cite{Friedman07,Rinaldo09} for its requirement of parameters with similar values, and overlapping group lasso with intra-group or inter-group sparsity \cite{Jacob09,Zhou10}. In fact, joint hyperedge and node selection formulation is a generalization of fused lasso to a the whole group (rather than only two variables for usual fused lasso).

\textbf{Out-of-Sample Extension.}
Previous models for learning on graphs and hypergraphs are only in transductive setting, requiring test samples to be included at training time. In our proposed formulations with star expansion in (\ref{starhyperedgesel}) (\ref{starnodesel2}) and (\ref{starjointsel}), we can have out of sample extension, meaning that labels of new samples can be predicted with $(\hat{\mu}_k,\delta_k), k=1,\dots,m$, without re-learning the model. This is advantageous compared to transductive learning models. For new samples (nodes) with known hyperedge membership, their labels $f$ can be inferred based on  its hyperedge representative/mean labels $\{\hat{\mu}_k\}_k$ and their smoothness measures $\delta_k$ by optimizing the same objective function as learning the model (\ref{genobj1}) with fixed $\mu_k$ as follows.
\begin{equation}
\hat{f} = \arg\min_{f} T_{e_k}(ss(f,e_k)) \ \ \ s.t \ \mu_k = \hat{\mu}_k.
\end{equation}

\subsection{Sparsistency}

We wish to see if these formulations would recover the support of a set of hyperedges that are relevant to the problem (model support consistency) in the same manner as lasso and others \cite{Zhao06,Wainwright09}. Since each hyperedge is associated with a smoothness measure $\delta_k$, a relevant hyperedge should have small enough $\delta_k$ and an irrelevant hyperedge should not. We first define relevant and irrelevant hyperedges based on their smoothness measures. We then introduce assumptions about noise level and how the hypergraph is constructed as $n \to \infty$ before main results showing the probability of recovering the correct sparse models. 

\begin{definition} \textbf{$\gamma$-smooth:}
Label function $f$  is considered $\gamma$-smooth on a given hyperedge $e$ if and only if its smoothness measure satisfies $ss(f,e) \leq \gamma$. 
\end{definition}

\begin{definition} \textbf{Relevant hyperedges:}
Given a $\gamma_r \in R$, a hyperedge $e$ is relevant to the problem with true label $y$ if and only if $y$ is $\gamma_r$-smooth on $e$. \end{definition}

\begin{definition} \textbf{Irrelevant hyperedges:}
Given a $\gamma_i \in R$, a hyperedge $e$ is irrelevant to the problem with true label $y$ if and only if $y$ is not $\gamma_i$-smooth on $e$.
\end{definition}

\subsubsection{General assumptions}
Below are the set of general assumptions behind our sparsistency analysis.

\textbf{(1) Noise assumption.}  Each observed label $Y_i$ is the true label $y_i$ plus white Gaussian noise: $(Y_i-y_i) \sim N(0,\delta^2)$.

\textbf{(2) Relevance assumption.}  We assume that all hyperedges in $E$ are either relevant ($\gamma_r$-smooth) or irrelevant (not $\gamma_i$-smooth) to the problem.

 \textbf{(3) Separation assumption.} $\gamma_i  > \gamma_r$ so that the two sets of hyperedges are disjoint. $\gamma_i  > \gamma_r$ should be large enough (depending on the models) to make sure that the two set of hyperedges are can be separated according to their smoothness measures with high probability.

Let $\gamma = \frac{\gamma_r+\gamma_i}{2}$ be the midpoint of $\{\gamma_r,\gamma_i\}$. Call the set of relevant hyperedges $E_r = \{e_1,\cdots e_r\}$ and irrelevant hyperedges $E_i = \{ e_{r+1}, \cdots e_m\}$. We call $E_r$ the \emph{support} of the problem. We now show the probability that  star expansion-styled joint hyperedge and node selection formulation (\ref{starjointsel})  and  star expansion-styled hyperedge selection  formulation   (\ref{starhyperedgesel}) will recover the correct (sparse) model support in the sense that the relevant hyperedges are $\gamma$-smooth but the irrelevant hyperedges are not.

\subsubsection{Sparsistency for hyperedge selection (\ref{starhyperedgesel})}

This is the case for $$ss = ss_2 = \min_{\mu_k \in R} w(e_k)\max_{i \in e_k} |f_i-\mu_k|.$$

This is equivalent to total variation in \cite{Hein13}. As also used in \cite{Hein13}, we let $w(e_k) = 1$ for the reason that smoothness measure does not increase linearly with the number of nodes in a hyperedge. 

To simplify discussion, we use the following \emph{growth model}. As having more training data $n \to \infty$, we assume that the hyperedge set is fixed and the number of points belonging to each hyperedge grows linearly with $n$, meaning that $r_k = \frac{|e_k|}{n} = const$ with respect to $n$. This makes $w(e_k) = \frac{1}{|e_k|} = O(n^{-1})$, and $n w(e_k)) = \frac{1}{r_k}$. Let $d \in R^n$ be the scaled weight vector: $d_i = n \sum_{k | i \in e_k} w(e_k)  = \sum_{k | i \in e_k} \frac{1}{r_k}$. Let $R = \max_k r_k$ and $D = \max_i \sum_{k| i \in e_k} d_k = \max_i \sum_{k| i \in e_k} \frac{1}{r_k}$, which are constants with respect to $n$. The sparsistency result is shown as follows.

\begin{theorem}\label{theo1}
Suppose that the general assumptions hold. Without loss of generality, assume that all labels have unit range $y_i,Y_i \in [0,1]$ and $\gamma_i - \gamma_r  > \frac{ 2\sqrt{2} \delta }{\sqrt{\pi}}$. As $ss(f,e_k) = \min_{\mu \in R}  \max_i |\mu-f_i|$,
let 
\begin{equation*}
\hat{f} = \arg\min_{f \in R^n} \frac{1}{2}|Y-f|^2 + \lambda \sum_k  ss(f,e_k) 
\end{equation*}
be the optimal solution of the learning problem (equivalent to (\ref{starhyperedgesel})). Then for $0< \lambda < \frac{\sqrt{\pi}(\gamma_i - \gamma_r)) - 2\sqrt{2}\delta}{D\sqrt{\pi}}$,   with a probability of at least $(1- \frac{1}{1+c^2n})^n$ for some constant $c$, it will recover the correct support of the problem in the sense that: 
\begin{enumerate}
\item $\hat{f}$ is $\gamma$-smooth on all hyperedges in $E_r$, and   
\item $\hat{f}$ is not $\gamma$-smooth on all hyperedges in $E_i$.
\end{enumerate}
\end{theorem}
Please find the detailed proof in the next Section. 

The condition $\gamma_i  - \gamma_r > 2\frac{\delta\sqrt{2}}{\sqrt{\pi}}$ is a necessary condition for sparsistency. This states that the smoothness gap $\gamma_i  - \gamma_r$ should be higher than the later term, which is the label error incurred by white noise in observed labels ($N(0,\delta^2)$). The condition for $\lambda$ means that it has to be small enough not to force the label to be smooth on all hyperedges. This looks counter-intuitive as one usually requires higher regularization hyperparameters to force \emph{irrelevant variables} to have zero weights in lasso \cite{Tibshirani96}. However, this actually makes sense by the nature of smooth functions in hypergraph situation where smaller $\lambda$ would not force all smoothness measures to be small, making smoothness measures of irrelevant hyperedges to be much larger than those of relevant hyperedges, resulting in a large gap and hence, sparsistency can be obtained.Similarly, $\lambda$ needs to be small enough to guarantee sparsistency. 

One thing to note about the probability is as follows.
$$\lim_{n \to \infty} (1- \frac{1}{1+c^2n})^n = 1- e^{\frac{-1}{c^2}} < 1$$
There is one \emph{key implication} is that,  the lower bound for probability of sparsistency $(1- \frac{1}{1+c^2n})^n$ does not converge to $1$ as $n \to \infty$. Sparsistency is only guaranteed with a limited probability regardless of sample size $n$.  This is a drawback of sparsistency result of total variation [9]. While this result does not automatically mean that sparsistency cannot be achieved, another argument can be used to show the limitation of hyperedge selection (total variation) as follows. 

\begin{proposition}
Suppose that $X_n$ is any set of $n$ white noises $x_i = Y_i-y_i  \sim N(0,\delta^2)$, then for any constant $a \in R$, 
$$
\lim_{n \to \infty} P( \max_{i \in X_n} |x_i| <a) = 0.  
$$
\end{proposition}
\begin{proof}
This is the case we assume that the differences between node labels and the representative label $\hat{\mu}_k$ follow a zero mean and fixed standard deviation. In this case, total variation, which is twice as large as the maximum absolute deviation, would go to infinity as $n \to \infty$ as 
\begin{align*}
\lim_{n \to \infty} P( \max_{i = 1 \cdots n} |x_i| <a) &= \lim_{n \to \infty}  P(|x_i| < a )  \forall i \\
 = \lim_{n \to \infty} P(|x|<a)^n &= 0.
\end{align*}
This is obvious as the larger the sample size is, the larger the maximum deviation (half of total variation) becomes. Hence, as $n \to \infty$, $\max_{i = 1 \cdots n}|Y_i - y_i|$ would go to infinity with high probability, exceeding the magnitude of the constants $\gamma_r$, $\gamma_i$ and $\delta$. Hence, at some large $n$ on, there is a high probability that $ss(f,e)$ is fully dominated by errors of the observed labels. This is different from the previous case of joint hyperedge and node selection where the mean of absolute deviation converges to a constant. This explains the lack of sparsistency guarantee in Theorem \ref{theo1}. 
\end{proof}

 \subsubsection{Sparsistency for joint hyperedge and node selection (\ref{starjointsel})} 
 
 This is the case for $$ss = ss_1 = \min_{\mu_k \in R} w(e_k)\sum_{i \in e_k} |f_i-\mu_k|.$$
 
  We assign $w(e_k) = \frac{1}{|e_k|}$, making   $ss(f,e_k) = \min_{\mu \in R}  \frac{1}{|e_k|} \sum_{i \in e_k} |f_i-\mu|$ the mean absolute deviation of labels on the hyperedge. This is suitable as we consider $n$ grows ($n \to \infty$) while keeping $\gamma_i$, $\gamma_r$ and $\delta$ constant. In this setting, all these parameters and $ss$ would be of the same magnitude. Note that there might be other weights, but in these cases, $\lambda$ cannot remain a constant, which not convenient for practical uses.

 We use the same growth model  with $n \to \infty$ and fixed hyperedge set, $\gamma_r$, $\gamma_i$ and $\delta$ to simplify discussion. 
  The following theorem shows the conditions and probability that the model recovers the correct support.

\begin{theorem}
Suppose that the general assumptions hold. Assume that the gap is wide enough with respect to label noise standard deviation $\delta$ in the sense that $\gamma_i  - \gamma_r > \frac{2\sqrt{2}\delta}{\sqrt{\pi}}$. As $ss(f,e_k) = \min_{\mu \in R}  \frac{1}{|e_k|} \sum_{i \in e_k} |f_i-\mu|$,
the optimal solution of the learning problem (equivalent to (\ref{starjointsel})) 
$$
\hat{f} = \arg\min_{f \in R^n} \frac{1}{2}|Y-f|^2 + \lambda \sum_k   ss(f,e_k) 
$$
for $0< \lambda <  \frac{ \sqrt{\pi}(\gamma_i  - \gamma_r) - 2\sqrt{2} \delta}{2\sqrt{\pi}D R}$ will, with a probability of at least $1- O(n^{-1})$, recover the correct support of the problem in the sense that 
\begin{enumerate}
\item $\hat{f}$ is $\gamma$-smooth on all hyperedges in $E_r$, and   
\item $\hat{f}$ is not $\gamma$-smooth on all hyperedges in $E_i$. 
\end{enumerate} 
\end{theorem}

Proof can be found in the next Section. The theorem states that under general assumptions, if the regularization parameter  ($\gamma$) is small enough, the optimization problem will find the correct sparsity pattern with a probability of at least $1- O(n^{-1})$. In other words, as $n \to \infty$, the probability of having sparsistency approaches 1 with a $O(n^{-1})$ rate. This means that the model is statistically consistent by converging to the true model with large enough  $n$.

The condition $\gamma_i  - \gamma_r > 2\frac{\delta\sqrt{2}}{\sqrt{\pi}} $ has the same interpretation as in Theorem \ref{theo1}. That is the level of noise should not exceed  the level of signal. The condition of mall enough $\lambda$ is also required.

All in all, we presented the conditions to recover the correct model support of the two formulations (\ref{starhyperedgesel}) and (\ref{starjointsel}) . While sparsistency of (\ref{starjointsel}) is guaranteed with large $n$, it is not the case for (\ref{starhyperedgesel}). 

We note that this is different from hyperedge lasso  \cite{Sharpnack12} in several aspects. First, our framework is strictly about learning smooth functions on a hypergraph. However, in \cite{Sharpnack12}, there is an additional constraint that labels from different hyperedges must be different. This is the case of non-overlapping hyperedges. As non-overlapping hyperedges divides the node set into disjoint subsets without any interrelationship, their labels can be learnt separately and this becomes unrelated problems.
Our setting of overlapping hyperedges is more realistic.

\section{Proof for Sparsistency}

Due to their lengths, we present sparsistency proofs in this section.

\subsection{Sparsistency for hyperedge selection} 

$$ss(f,e_k) = \min_{m_k \in R} w(e_k) \max_i |m_k-f_i|$$

We first prove two lemmas to support the main proof. For $w(e_k) = 1$.

\begin{lemma}\label{randombound2}
 Given a true label function $y$ on a hypergraph. Suppose that label $Y$ is a noisy observation with white noise in the sense that $(y_i - Y_i)$ are independent following a normal distribution $N(0,\delta^2)$. Then, with a probability of at least $\frac{t^{2n}}{(1+t^{2})^n}$, $max |y_i-Y_i| < \frac{\delta \sqrt{2}}{\sqrt{\pi}} + t \delta \frac{\sqrt{\pi -2}}{\sqrt{n\pi}}$  and $\forall t \geq 0$. 
 \end{lemma}
 
\begin{proof}

As $(y_i - Y_i)$ follows normal distribution with standard deviation $\delta$, according to one-sided Chebyshev's inequality (also known as Cantelli's lemma), for any $i$,
$$
P(|y_i - Y_i| - \frac{\delta \sqrt{2}}{\sqrt{\pi}} \leq t \delta \frac{\sqrt{\pi -2}}{\sqrt{n\pi}}) \geq  \frac{t^2}{1+t^2}.
$$
Then, $\forall i = 1..n$,  
\begin{align*}
&P(\max_i |y_i - Y_i| -  \frac{\delta \sqrt{2}}{\sqrt{\pi}}  \leq t \delta \frac{\sqrt{\pi -2}}{\sqrt{n\pi}}) 
\\ 
= &P(\forall i: |y_i - Y_i| - \frac{\delta \sqrt{2}}{\sqrt{\pi}} \leq t \delta \frac{\sqrt{\pi -2}}{\sqrt{n\pi}}) \\
\geq & \frac{t^{2n}}{(1+t^{2})^n}.
\end{align*}
\end{proof}

\begin{lemma}\label{lbound2}
 Given any label $Y$ on a hyperedge $e$.  Let $\max  |f-Y| = \max_{i \in 1..n} |f_i-Y_i|$ and likewise $\max |f-Y|_e = \max_{i \in e} |f_i-Y_i|$.  
 Any close approximate $f$ of $Y$ would have smoothness measure close enough to that of $Y$ by: 
\begin{align}
&ss(Y,e) - w(e)\max |f-Y|_e \nonumber\\
\leq & ss(f,e) \nonumber\\
\leq & ss(Y,e) + w(e)\max |f-Y|_e.
\end{align}
\end{lemma}

\begin{proof}

Let  $\mu_y = \arg \min_{\mu \in R} w(e)\sum_{i \in e} |y_i-\mu|$ and $\mu_f = \arg \min_{\mu \in R} w(e)\sum_{i \in e} |f_i-\mu|$. We prove the second inequality.
\begin{align*}
ss(f,e) &=  w(e_k)\max_{i \in e} |f_i-\mu_f| \\
& \leq w(e)\max_{i \in e} |f_i-\mu_y| \\
& \leq  w(e)\max_{i \in e} (  |Y_i-\mu_y| + |f_i-Y_i|) \\
& \leq ss(Y,e) + w(e) \max |f-Y|_e. \\
\end{align*} 
Due to symmetry, we can also have the first inequality.
\begin{align*}
ss(Y,e)  &\leq ss(f,e) + w(e) \max |f-Y|_e\\
ss(Y,e) - w(e)\max |f-Y|_e &\leq ss(f,e).
\end{align*}

\end{proof}

We now restate and prove the main theorem

\textbf{Theorem 1.} \textit{Suppose that the general assumptions hold. Without loss of generality, assume that all labels have unit range $y_i,Y_i \in [0,1]$ and $\gamma_i - \gamma_r  > \frac{ 2\sqrt{2} \delta }{\sqrt{\pi}}$. As $ss(f,e_k) = \min_{\mu \in R}  \max_i |\mu-f_i|$,
let 
\begin{equation*}
\hat{f} = \arg\min_{f \in R^n} \frac{1}{2}|Y-f|^2 + \lambda \sum_k  ss(f,e_k) 
\end{equation*}
be the optimal solution of the learning problem (equivalent to (7)). Then for $0< \lambda < \frac{\sqrt{\pi}(\gamma_i - \gamma_r)) - 2\sqrt{2}\delta}{D\sqrt{\pi}}$,   with a probability of at least $(1- \frac{1}{1+c^2n})^n$ for some constant $c$,  it will recover the correct support of the problem in the sense that: 
\begin{enumerate}
\item $\hat{f}$ is $\gamma$-smooth on all hyperedges in $E_r$, and   
\item $\hat{f}$ is not $\gamma$-smooth on all hyperedges in $E_i$.
\end{enumerate}
}

\begin{proof}

\textbf{Step 1:} Bounding $\max |f-Y|$.

For $0 \leq Y_i \leq 1$, let $m_k = \arg\min_m \max_{i \in e_k} (f_i-m)$. Hence $m_k$ becomes the mid-range value of $f_i$, and therefore $0 \leq m_k \leq 1$.

Let $f$ (abuse the notation of $\hat{f}$ for simplicity) be the optimal solution of $f = \arg\min_{f \in R^n} \frac{1}{2}|Y-f|^2 + \lambda \sum_k  ss(f,e_k)$. Then, at the optimal point, 
$$
\frac{\partial(
\frac{1}{2}|Y-f|^2 +  \lambda \sum_k  ss(f,e_k))}{\partial f} = 0. 
$$
Set the partial derivative on $f_i$ to be zero. Derivative of the first term (quadratic) is $f-Y$.
Note that $ss(f,e_k)$ is not differentiable. However, we can have the subderivatives of $\sum_k ss(f,e_k)$ and $-(f-Y)$ should be in this range. 

We show that any directional derivative of $\sum_k ss(f,e_k)$ for $f_i$ is contained within $[-\lambda \frac{d_i}{2}, \lambda \frac{d_i}{2}]$.

\textbf{Step 1.1:} For each $e_k$.   Let take an infinitesimal change $\partial f_i $ of $f_i$. If $f_i$ is not at the border ($|f_i-\mu_k| < max_j |f_j-\mu_k| = ss(f,e_k)$), then $f_i + \partial f_i$ would not change $ss(f,e_k)$, making the directional derivative $0$. If $f_i$ is at the border  ($|f_i-\mu_k| = max_j |f_j-\mu_k| = \frac{ss(f,e_k)}{w(e_k)}$) then taking an infinitesimal change $\partial f_i$ of $f_i$ still keeps $f_i$ at the border, and $\mu_k$, being the mid-range, would change $\frac{\partial f_i}{2}$. In this case, $ss(f,e_k)$ would change $\frac{w(e_k) \partial f_i}{2}$ making the directional derivative  either $\frac{w(e_k)}{2}$ or $\frac{-w(e_k)}{2}$. Then the directional derivative of $ss(f,e_k)$ on $f_i$ is bounded by $\frac{\partial ss(f,e_k)}{\partial f_i}  \in [-\frac{\lambda w(e_k)}{2}, \frac{\lambda w(e_k)}{2}]$.

\textbf{Step 1.2:}  For the whole hypergraph: summing from all $e_k$, we have upper bound
\begin{align*}
(f_i-Y_i) \in &[ -\lambda \frac{ \sum_{k | i \in e_k} w(e_k)}{2}, \lambda \frac{ \sum_{k | i \in e_k} w(e_k)}{2}] \\
= &[- \lambda \frac{d_i}{2}, \lambda \frac{d_i}{2}]
\end{align*}
 ($-(f_i-Y_i)$ is in the range of the subderivative of the latter term).
 
Therefore, we have:
\begin{align}\label{estbound2}
&- \frac{\lambda d}{2} \leq \hat{f} - Y \leq \frac{\lambda d}{2}  \nonumber \\
& max |f -Y| < max_i  \frac{\lambda d_i}{2} = \frac{\lambda D}{2}
\end{align}
(as $D = max_i   d_i$ being the max degree of any node, being constant with respect to $n$).

\textbf{Step 2:} bounding $ss(f,e_k)$.

\textbf{Step 2.1:} For relevant hyperedges $E_r$: $k=1..r$:
\begin{itemize}
\item $ss(y,e_k) < \gamma_r$ as in the general assumptions.
\item $ss(Y,e_k) < \gamma_r + w(e_k)\max |y-Y|_{e_k}$ (Lemma \ref{lbound2}). 
\end{itemize}
Then, according to Lemma \ref{randombound2}, for any $t >0$, with a probability of at least $\frac{t^{2n}}{(1+t^{2})^n}$, we can have:
\begin{equation}
ss(Y,e_k) < \gamma_r + \frac{\delta \sqrt{2}}{\sqrt{\pi}} + t \delta \frac{\sqrt{\pi -2}}{\sqrt{n\pi}}. 
\end{equation}

Then, according to Lemma \ref{lbound2} and (\ref{estbound2}), 
\begin{equation}
ss(f,e_k) < \gamma_r + \frac{\delta \sqrt{2}}{\sqrt{\pi}} + t \delta \frac{\sqrt{\pi -2}}{\sqrt{n\pi}} + \frac{\lambda D}{2}
\end{equation}
with a probability of at least $\frac{t^{2n}}{(1+t^{2})^n}$.

\textbf{Step 2.2:} For irrelevant hyperedges $E_i$: $k= (r+1)..m$
\begin{itemize}
\item $ss(y,e_k) > \gamma_i$ as in the general assumptions.
\item $ss(Y,e_k)> \gamma_i - w(e_k)\max |y-Y|_{e_k}$ (Lemma \ref{lbound2}). 
\end{itemize}

Then, according to Lemma \ref{randombound2}, for any $t >0$, with a probability of at least $\frac{t^{2n}}{(1+t^{2})^n}$, we can have:
\begin{equation}\label{bound3}
ss(Y,e_k) > \gamma_i - \frac{\delta \sqrt{2}}{\sqrt{\pi}} - t \delta \frac{\sqrt{\pi -2}}{\sqrt{n\pi}}. 
\end{equation}

Then, according to Lemma \ref{lbound2} and (\ref{estbound2}), 

\begin{equation}\label{bound4}
ss(f,e_k) > \gamma_i - \frac{\delta \sqrt{2}}{\sqrt{\pi}} - t \delta \frac{\sqrt{\pi -2}}{\sqrt{n\pi}} - \frac{\lambda D}{2}
\end{equation}
with a probability of at least $\frac{t^{2n}}{(1+t^{2})^n}$.

Hence, from (\ref{bound3}) and (\ref{bound4}) we can bound the estimated $f$.

\textbf{Step 2.3:} For all hyperedges.  We investigate the condition that there exists $\lambda > 0$ and $t>0$ such that smoothness measures of relevant hyperedges are smaller than those of irrelevant hyperedges:
 $\gamma_r + \frac{\delta \sqrt{2}}{\sqrt{\pi}} + t \delta \frac{\sqrt{\pi -2}}{\sqrt{n\pi}} + \frac{\lambda D}{2} < \gamma$ $\forall k = 1..r$ (equivalently, $ \gamma_i - \frac{\delta \sqrt{2}}{\sqrt{\pi}} - t \delta \frac{\sqrt{\pi -2}}{\sqrt{n\pi}} - \frac{\lambda D}{2} > \gamma$ $\forall k = (r+1)..m$). This is equivalent to $\gamma_r + \frac{\delta \sqrt{2}}{\sqrt{\pi}} < \gamma$, or 
\begin{equation}\label{cond2}
\gamma_i - \gamma_r  > \frac{ 2\sqrt{2} \delta }{\sqrt{\pi}}.
\end{equation}

With the condition  that $\gamma_r + \frac{\delta \sqrt{2}}{\sqrt{\pi}} < \gamma$, then we can choose $\lambda >0$ such that $\gamma - \gamma_r - \frac{\delta \sqrt{2}}{\sqrt{\pi}} - \frac{\lambda D}{2} > 0$ $\forall k=1..r$, equivalently $\gamma_i -\gamma - \frac{\delta \sqrt{2}}{\sqrt{\pi}}- \frac{\lambda D}{2}  >0$  $\forall i= (r+1)..m$), or equivalently:

 \begin{align}\label{lambdacond2}
 \lambda &< \frac{2(\gamma - \gamma_r)}{D } - \frac{2\sqrt{2}\delta}{D \sqrt{\pi}} = \frac{\sqrt{\pi}(\gamma_i - \gamma_r)) - 2\sqrt{2}\delta}{D\sqrt{\pi} }.
\end{align}

With $\lambda$ satisfying condition (\ref{lambdacond2}), then we choose $t$ so that $\gamma_r + \frac{\delta \sqrt{2}}{\sqrt{\pi}} + t \delta \frac{\sqrt{\pi -2}}{\sqrt{n\pi}} + \frac{\lambda D}{2} < \gamma$ (equivalently, $ \gamma_i - \frac{\delta \sqrt{2}}{\sqrt{\pi}} - t \delta \frac{\sqrt{\pi -2}}{\sqrt{n\pi}} - \frac{\lambda D}{2} > \gamma$). Choosing the largest possible $t_0$: 
\begin{equation}\label{t2} 
t_0 = \frac{\sqrt{n\pi}}{\delta \sqrt{\pi-2}} (\gamma - \gamma_r - \frac{\delta \sqrt{2}}{\sqrt{\pi}}  - \frac{\lambda D}{2}).
\end{equation}
Note that $t_0 = O(\sqrt{n})$. To simplify \ref{t2}, let $t = c \sqrt{n}$ for some constant $c$. For $\lambda$ satisfying (\ref{lambdacond2}), $t_0$ satisfying (\ref{t2}) we can conclude that 
\begin{itemize}
\item $ss(f,e_k) < \gamma$ $\forall k = 1..r$ and
\item  $ss(f,e_k) > \gamma$ $\forall k = (r+1)..m$
\end{itemize}
with a probability of at least $\frac{t_0^{2n}}{(1+t_0^{2})^n} = (1- \frac{1}{1+c^2n})^n$. 

Note that, $\lim_{n \to \infty (1- \frac{1}{1+c^2n})^n = (1- \frac{1}{c^2n})^n} = e^{\frac{-1}{c^2}}$ (with $e$ being  Euler's number, the base of natural logarithm) not converging to 1.

\end{proof}

\subsection{Sparsistency for joint selection} 
 
 $$ss(f,e_k) = \min_{\mu \in R} w(e_k) \sum_{i \in e_k} |f_i-\mu_k|$$

 We set $w(e) = \frac{1}{|e|}$ to avoid hyperedge weight size effect. This setting makes the smoothness measure on each hyperedge become the average variance.  Let $r_k  = \frac{|e_k|}{n}$, $R = \max_{k = 1..m} r_k$ and $R^{'} = \min_{k = 1..m} r_k$. For simplicity, we assume that the number of nodes in a hyperedge grow linearly with hypergraph size, meaning that $r_k$ are constant, so are $R$ and $R^{'}$. Hence $w(e_k) = \frac{1}{nr_k} = O(n^{-1})$. 
 
 We first prove some lemmas to support the main proof.
 
 \begin{lemma}\label{randombound1}
 Given a true label function $y$ on a hypergraph. Suppose that label $Y$ is a noisy observation with white noise in the sense that $(y_i - Y_i)$ are independent following a normal distribution $N(0,\delta^2)$. Then, with a probability of at least $1- \frac{1}{1+t^2}$,  for $\forall t \geq 0$, $\Big( \frac{|y-Y|_1}{n} - \frac{\delta\sqrt{2}}{\sqrt{\pi}} \Big)  \leq t\delta \sqrt{\frac{\pi - 2}{\pi n}}$. 
 \end{lemma}
 
\begin{proof}
As $(y_i - Y_i)$ is follows normal distribution with standard deviation $\delta$, $|y_i - Y_i|$ follows a half-normal distribution with a mean of $\frac{\delta\sqrt{2}}{\sqrt{\pi}}$  and a standard deviation of $\delta\sqrt{1-\frac{2}{\pi}}$.

Hence, $|y-Y|_1 = \sum_{i=1}^n |y_i-Y_i|$, as the sum of $n$ independent random variables, will have a mean of $\frac{n\delta\sqrt{2}}{\sqrt{\pi}}$ and a standard variation of $\delta\sqrt{n(1-\frac{2}{\pi})}$
 
 According to one-sided Chebyshev's inequality (also known as Cantelli's inequality), 
 $$
 P \bigg( \Big( |y-Y|_1 - \frac{n\delta\sqrt{2}}{\sqrt{\pi}} \Big)  \leq t\delta \sqrt{n(1-\frac{2}{\pi})} \bigg) \geq \frac{t^2}{1+t^2}
 $$
 Dividing the inequality by $n$, then 
 $$
 P \bigg( \Big( \frac{|y-Y|_1}{n} - \frac{\delta\sqrt{2}}{\sqrt{\pi}} \Big)  \leq t\delta \sqrt{\frac{\pi-2}{\pi n}} \bigg) \geq \frac{t^2}{1+t^2}
 $$
 meaning that the probability that $\frac{|y-Y|_1}{n}$ is far enough from its expected value $\delta\frac{\sqrt{2}}{\sqrt{\pi}}$ approaches zero as $n \to \infty$. 
\end{proof}

 \begin{lemma}\label{lbound1}
 Given any two labels $f$ and $Y$ on a hyperedge $e$.  Let $|f-Y|_e = \sum_{i \in e} |f_i-Y_i|$.
 Then their smoothness measure on $e$ is  bounded by each other in the sense that. 
\begin{equation}
ss(Y,e) - w(e)|f-Y|_e \leq ss(f,e) \leq ss(Y,e) + w(e)|f-Y|_e.
\end{equation}
\end{lemma}

\begin{proof}

Let  $\mu_y = \arg \min_{\mu \in R} w(e)\sum_{i \in e} |y_i-\mu|$ and $\mu_f = \arg \min_{\mu \in R} w(e)\sum_{i \in e} |f_i-\mu|$.

\begin{align*}
ss(f,e) &=  \min_{\mu \in R} w(e)\sum_{i \in e} |f_i-\mu| \\
& = w(e)\sum_{i \in e} |f_i-\mu_f| \\
& \leq w(e)\sum_{i \in e} |f_i-\mu_y| \\
& \leq  w(e)\sum_{i \in e} (  |Y_i-\mu_y| + |f_i-Y_i|) \\
& = w(e) \sum_{i \in e_k}   |Y_i-\mu_k| + w(e)|f-Y|_e. 
\end{align*} 
This proves the second inequality. Due to symmetry, the first inequality can be equivalently induced.
\end{proof}

We now restate and prove the main theorem.

\textbf{Theorem 2.} \textit{
Suppose that the general assumptions hold. Assume that the gap is wide enough with respect to label noise standard deviation $\delta$ in the sense that $\gamma_i  - \gamma_r > \frac{2\sqrt{2}\delta}{\sqrt{\pi}}$. As $ss(f,e_k) = \min_{\mu \in R}  \frac{1}{|e_k|} \sum_{i \in e_k} |f_i-\mu|$,
the optimal solution of the learning problem (equivalent to (9)) 
$$
\hat{f} = \arg\min_{f \in R^n} \frac{1}{2}|Y-f|^2 + \lambda \sum_k   ss(f,e_k) 
$$
for $0< \lambda <  \frac{ \sqrt{\pi}(\gamma_i  - \gamma_r) - 2\sqrt{2} \delta}{2\sqrt{\pi}D R}$ will, with a probability of at least $1- O(n^{-1})$, recover the correct support of the problem in the sense that 
\begin{enumerate}
\item $\hat{f}$ is $\gamma$-smooth on all hyperedges in $E_r$, and   
\item $\hat{f}$ is not $\gamma$-smooth on all hyperedges in $E_i$.
\end{enumerate}
}

The theorem states that under general conditions, if the regularization parameter is small enough, the optimization problem will find the right ($\gamma$) sparsity pattern with a probability that approaches 1 as $n \to \infty$. We abuse the notation to use $f_i$ for $\hat{f}_i$ for simplicity from here on.

\begin{proof}

\textbf{Step 1:} bounding $f-Y$.

As  $\hat{f} = \arg\min_{f \in R^n} \frac{1}{2}|Y-f|^2 + \lambda \sum_k ss(f,e_k)$, at the optimal point, derivative of the objective function becomes zero.
$$
\frac{\partial(
\frac{1}{2}|Y-f|^2 +  \lambda \sum_k ss(f,e_k))}{\partial f} = 0. \\
$$
Set the partial derivative on each $f_i$ to be zero. Derivative of the first term (quadratic) is $(f_i-Y_i)$. The second term is not everywhere differentiable. However, we can have the subderivative of $\sum_k ss(f,e_k)$ and $-(f_i-Y_i)$ in its range at optimal point of $f$. 

\textbf{Step 1.1:}  For each $e_k$, we show that any directional derivative of $ss(f,e_k)$ is contained within $[-\lambda w(e_k), \lambda w(e_k)]$.  Note that as $ \mu_k = \arg\min_{\mu \in R} w(e_k) \sum_{i \in e_k} |f_i-\mu|$, $\mu_k$ is either the median of all $\{f_i\}_{i \in e_k}$ if $|e_k|$ is an odd number or within the two medians of $\{f_i\}_{i \in e_k}$ if $|e_k|$ is an even number. 

Let take an infinitesimal change $\partial  f_i$ of $f_i$, then $\mu_k$ only changes if $i$ is exactly the median point ($f_i = \mu_k$). In this case, the total change to $ss(f,e_k)$ is $0$ as $\mu_k$ changes with $f_i$, $f_i-\mu_k = 0$. For all the other cases,  $ss(f,e_k)$  only changes at $|f_i-\mu_k|$ being $w(e_k)\partial f_i$ as $\mu_k$ remains the same. 
Hence, the maximum change of $ss(f,e_k)$ is $w(e_k)\partial f_i$. Then, the directional derivative of $ss(f,e_k)$ on $f_i$ is bounded by $\frac{\partial ss(f,e_k)}{\partial f_i} \in [-\lambda w(e_k), \lambda w(e_k)]$. 

\textbf{Step 1.2:} For the whole hypergraph, summing up from all $e_k$: 

$$(f_i-Y_i) \in [ -\lambda \sum_{k| i \in e_k}w(e_k) , \lambda\sum_{k| i \in e_k} w(e_k)].$$

 Then $d$ is constant with respect to $n$.  Then we have
\begin{equation}\label{estbound1}
\frac{-\lambda d}{n} \leq f- Y \leq \frac{\lambda d}{n}.
\end{equation}

\textbf{Step 2:} bounding $ss(f,e_k)$.

\textbf{Step 2.1:}  For relevant hyperedges in $E_r$: $k= 1..r$
\begin{itemize}
\item $ss(y,e_k) \leq \gamma_r$ as in the general assumptions.
\item $ss(Y,e_k) \leq  ss(y,e_k)+ w(e_k) |y-Y|_{e_k}$ (Lemma \ref{lbound1}).
\end{itemize}
Then, according to Lemma \ref{randombound1}, for any $t>0$, with a probability of at least $1-\frac{1}{1+t^2}$, 
\begin{align}
ss(Y,e_k) &< \gamma_r +  \frac{\delta\sqrt{2}}{\sqrt{\pi}} +  t\delta \sqrt{\frac{\pi - 2}{\pi |e_k|}} \nonumber\\
&= \gamma_r +  \frac{\delta\sqrt{2}}{\sqrt{\pi}} +   t\delta \frac{\sqrt{\pi - 2}}{\sqrt{n r_k \pi}}.
\end{align}

As $D = \max_i \sum_{k| i \in e_k} d_k = \max_i \sum_{k| i \in e_k} \frac{1}{r_k}$, lemma \ref{lbound1} together with (\ref{estbound1}) gives:
\begin{align}\label{bound1}
ss(f,e_k) &< \gamma_r + \frac{\delta\sqrt{2}}{\sqrt{\pi}} +   t\delta \frac{\sqrt{\pi - 2}}{\sqrt{n r_k \pi}} +   \sum_{i \in e_k} \frac{\lambda d_i}{n} \nonumber\\
&\leq \gamma_r + \frac{\delta\sqrt{2}}{\sqrt{\pi}} +   t\delta \frac{\sqrt{\pi - 2}}{\sqrt{n r_k \pi}} + \lambda D r_k.
\end{align}

\textbf{Step 2.2:}  For irrelevant hyperedges in $E_i$, $k= (r+1)..m$
\begin{itemize}
\item $ss(y,e_k) > \gamma_i$ as in the general assumptions.
\item $ss(Y,e_k) >  ss(y,e_k)- w(e_k) |y-Y|_{e_k}$ (Lemma \ref{lbound1}) 
\end{itemize}

Then, according to Lemma \ref{randombound1}, for any $t>0$, with a probability of at least $1-\frac{1}{1+t^2}$, 
\begin{align}
ss(Y,e_k) &> \gamma_i -  \frac{\delta\sqrt{2}}{\sqrt{\pi}} -  t\delta \sqrt{\frac{\pi - 2}{\pi |e_k|}} \nonumber\\
&= \gamma_i -  \frac{\delta\sqrt{2}}{\sqrt{\pi}} -   t\delta \frac{\sqrt{\pi - 2}}{\sqrt{n r_k \pi}}.
\end{align}

Lemma \ref{lbound1} together with (\ref{estbound1}) gives:
\begin{align}\label{bound2}
ss(f,e_k) &> \gamma_i - \frac{\delta\sqrt{2}}{\sqrt{\pi}} -   t\delta \frac{\sqrt{\pi - 2}}{\sqrt{n r_k \pi}} -\sum_{i \in e_k} \frac{\lambda d_i}{n} \nonumber\\
&> \gamma_i - \frac{\delta\sqrt{2}}{\sqrt{\pi}} -  t\delta \frac{\sqrt{\pi - 2}}{\sqrt{n r_k \pi}} - \lambda D r_k.
\end{align}
Hence, from (\ref{bound1}) and (\ref{bound2}), we can bound the estimated $f$.  

\textbf{Step 2.3:} For all hyperedges. We investigate the condition that there exist $\lambda >0 $ and $t>0$ so that smoothness measures of relevant hyperedges are smaller than those of irrelevant hyperedges: $\gamma_r + \frac{\delta\sqrt{2}}{\sqrt{\pi}} < \gamma $, then for $\lambda > 0$ such that $\gamma  - \gamma_r - \frac{\delta\sqrt{2}}{\sqrt{\pi}}  - \lambda D r_k > 0$ $\forall k=1..r$, (equivalently  $\gamma_i - \frac{\delta\sqrt{2}}{\sqrt{\pi}}  - \lambda D R - \gamma > 0$ $\forall k=(r+1)..m$ ), or equivalently:
\begin{equation}\label{lambda1}
\lambda < \frac{\gamma  - \gamma_r - \frac{\delta\sqrt{2}}{\sqrt{\pi}}}{D R} = \frac{ \sqrt{\pi}(\gamma_i  - \gamma_r) - \sqrt{2} \delta}{\sqrt{\pi}D R}
\end{equation}
as $R = \max_k r_k$ in the growth model. 

For $\lambda$ satisfying this condition (\ref{lambda1}), then for any $t$ so that $\gamma_r + \frac{\delta\sqrt{2}}{\sqrt{\pi}} +   t\delta \frac{\sqrt{\pi - 2}}{\sqrt{n r_k \pi}} + \lambda D R \leq \gamma$  $\forall k$ (equivalently $\gamma_i - \frac{\delta\sqrt{2}}{\sqrt{\pi}} -  t\delta \frac{\sqrt{\pi - 2}}{\sqrt{n r_k \pi}} - \lambda D r_k  \geq \gamma$  $\forall k$), such as: 

 \begin{equation}\label{t1}
t_0 = (\gamma - \gamma_r -  \frac{\delta\sqrt{2}}{\sqrt{\pi}}  -  \lambda DR)\frac{\sqrt{n R^{'} \pi }}{\delta\sqrt{\pi-2}} 
\end{equation}

Note that $ t_0 = O(\sqrt{n})$. For $\lambda$ satisfying condition in (\ref{lambda1}), $t_0$ satisfying the condition in (\ref{t1}), we can finally conclude that
\begin{itemize}
\item $ss(f,e_k) < \gamma$ $\forall k = 1..r$, and 
\item $ss(f,e_k) > \gamma$ $\forall k = (r+1)..m$
\end{itemize}
with a probability of at least $1- \frac{1}{1+t_0^2} =1 - O(n^{-1})$.
\end{proof}

\section{Simulation}
We designed experiments with simulated data to demonstrate how sparsely smooth models work and the benefit of sparsity. Hypergraphs were generated as follows. Labels vector $Y$ for 200 nodes were i.i.d. samples following a uniform distribution on $[0,1]$. Ten \emph{relevant hyperedges} were generated  as ten groups of nodes $i$ that $a \leq y(i) < a+0.15$ for evenly spaced $a$ from 0 to 0.9. A number  of  \emph{irrelevant hyperedges} (depending on the simulations) were generated by randomly grouping 20 points regardless of $Y$. A number of \emph{noisy nodes} were randomly chosen to add into  each relevant hyperedge. Ten-fold cross-validation to split train/test data and averaged RMSEs over ten runs. We used transductive setting to include test data at learning stage for all models. The lower the RMSE, the better the model. Tradeoff parameter $\lambda$ was chosen to be the one with the lowest RMSE among $\{10^{i-5}\}, i=1..7$. 
To show the properties of different sparsely smooth models, we compare the following models: 
\begin{itemize}
\item \textbf{(i) dense:} the non-sparse model in the sense that it does not use $l_1$ regularization (equivalent to star and clique expansions  \cite{Zhou06,Agarwal06} and then using graph Laplacian regularization), 
\item \textbf{(ii) hyperedge selection:} (\ref{starhyperedgesel}) equivalent to total variation model \cite{Hein13}, 
\item \textbf{(iv) node selection:} (\ref{starnodesel2}) (our proposed formulation),
\item \textbf{(iii) joint selection:} (\ref{starjointsel}) (our proposed formulation).
\end{itemize}

\subsection{Irrelevant Hyperedges}
\begin{figure}[t]
  	\includegraphics[width=1\linewidth]{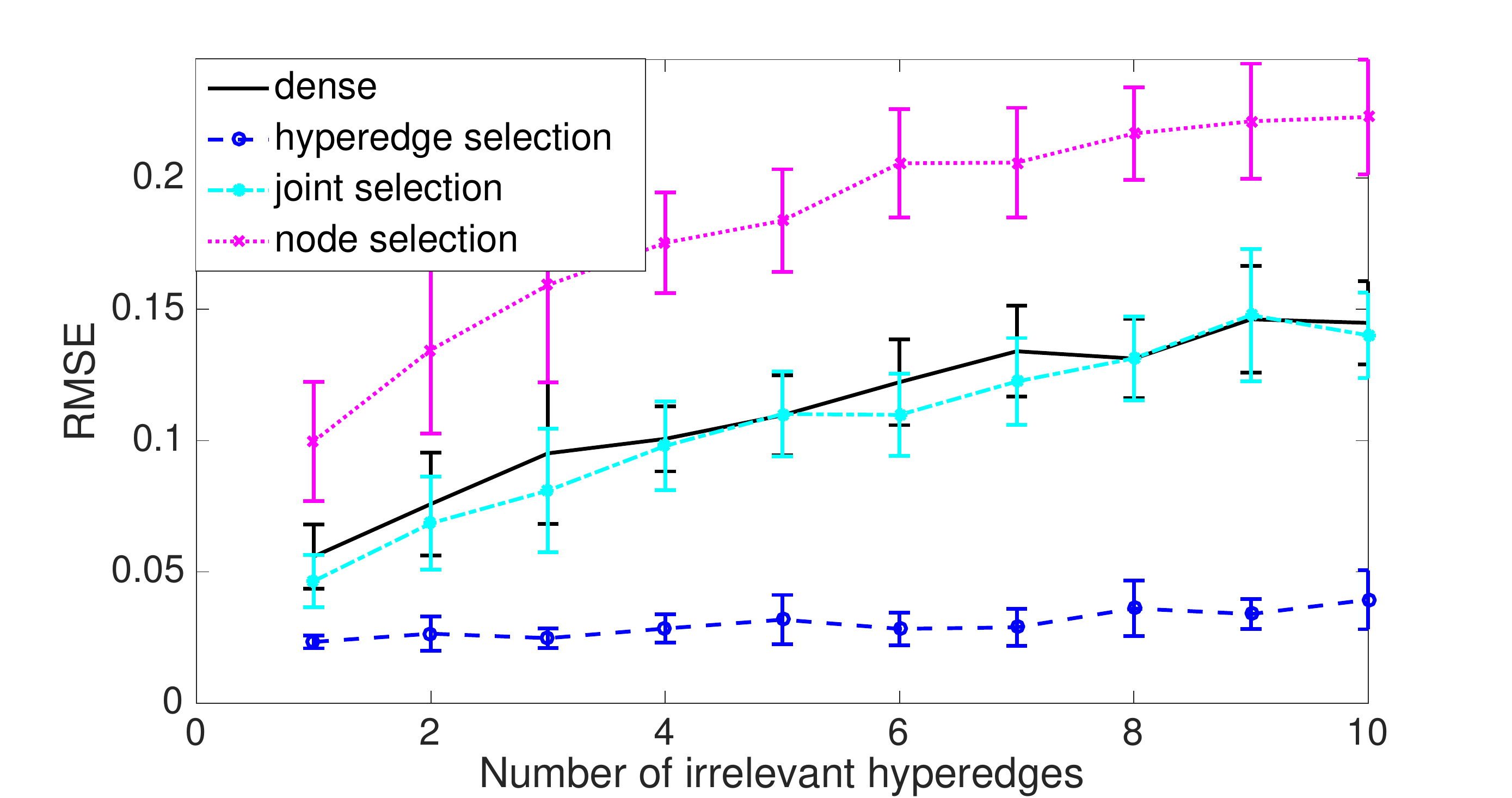}
 	\caption{Simulation  with  irrelevant hyperedges without noisy nodes. X axis was the number of irrelevant hyperedges and y axis was the RMSEs of different models on test data. Hyperedge selection method gave the lowest RMSEs consistently at different numbers of irrelevant hyperedges.}
  	\label{fig:sim1}
    \end{figure}

This is to simulate the situation that there are irrelevant hyperedges in data. We generated hypergraphs with different numbers of irrelevant hyperedges ranging from 1 to 10, without noisy nodes. We split the data to train and test as described above.
We summarized RMSEs of the models for different numbers of irrelevant hyperedges in Figure \ref{fig:sim1}. We could observe that hyperedge selection model had very stable performances even when the number of irrelevant hyperedges increased. In contrast, the other models were sensitive to the number of irrelevant hyperedges. This showed that
 hyperedge selection model could select relevant hyperedges  in presence of irrelevant hyperedges without noisy nodes while the other methods were not as effectively.

\subsection{Noisy Nodes}
\begin{figure}[t]
         \includegraphics[width=1\linewidth]{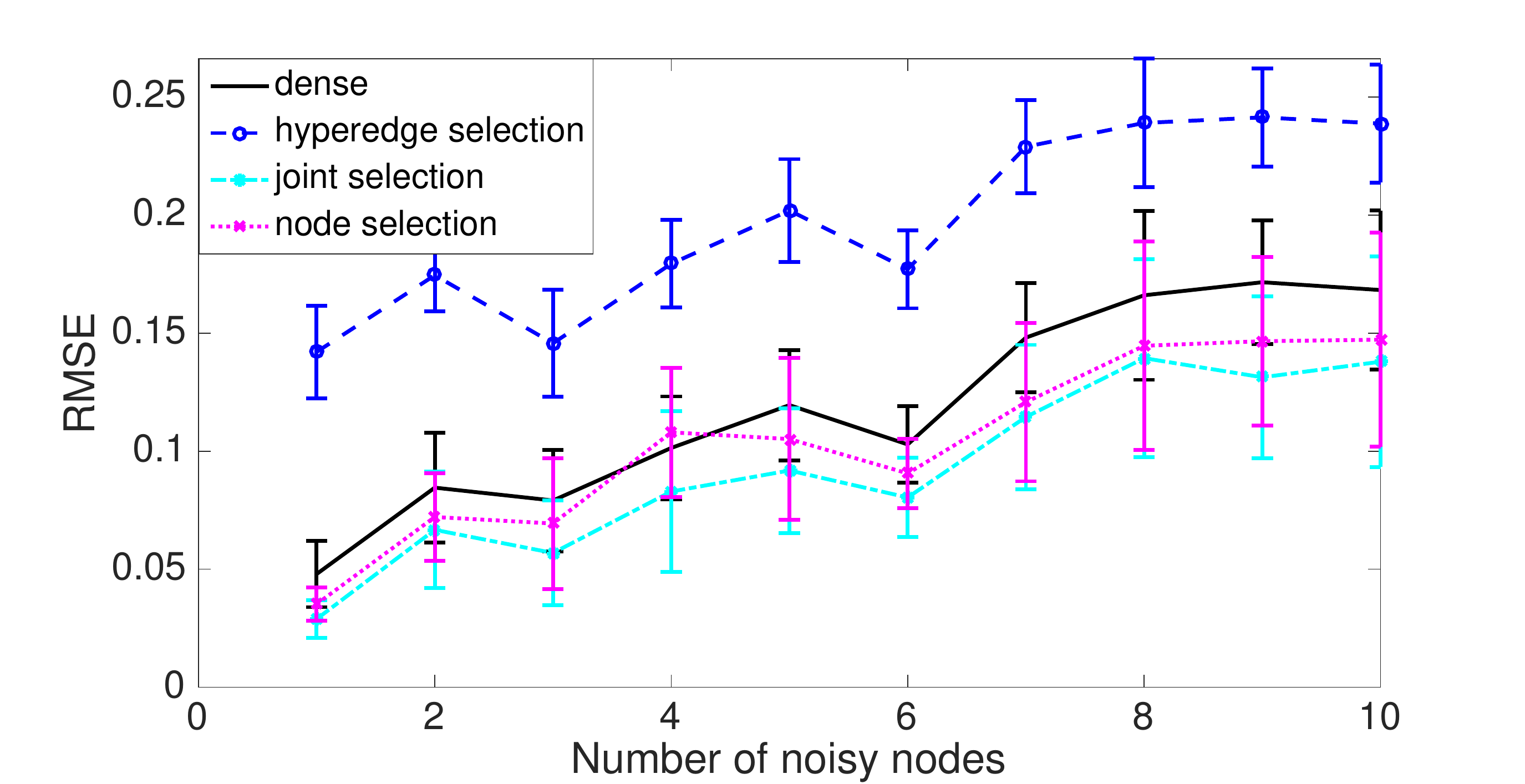}
	 \caption{Simulation with  noisy nodes and without irrelevant hyperedges. X axis was the number of noisy nodes added to each hyperedge and y axis was the RMSEs of the models. Hyperedge selection model had the worst performances while the node selection and joint selection models performed best in terms of RMSEs.}
  	\label{fig:sim3}
    \end{figure}

We simulated the case of having only noisy nodes inside relevant hyperedges (no irrelevant hyperedges). We generated hypergraphs with different noisy nodes and without irrelevant hyperedges. We slitted train/test data and conducted experiments in the same way  as above.  We showed RMSEs of the models for different numbers of noisy nodes in Figure \ref{fig:sim3}.  We found that hyperedge selection model has the worst performance in terms of RMSEs consistently. The other models, while sensitive to the number of noisy nodes, have lower better performances with lower RMSEs.  Joint hyperedge and node selection model were the one with the lowest RMSEs, then followed by node selection and  dense models. The experiments showed that joint  selection model had similar performances to node selection models only, showing its flexibility in case of noisy nodes without irrelevant hyperedges.

\subsection{Both Irrelevant Hyperedges and Noisy Nodes}
\begin{figure}[t]
         \includegraphics[width=1\linewidth]{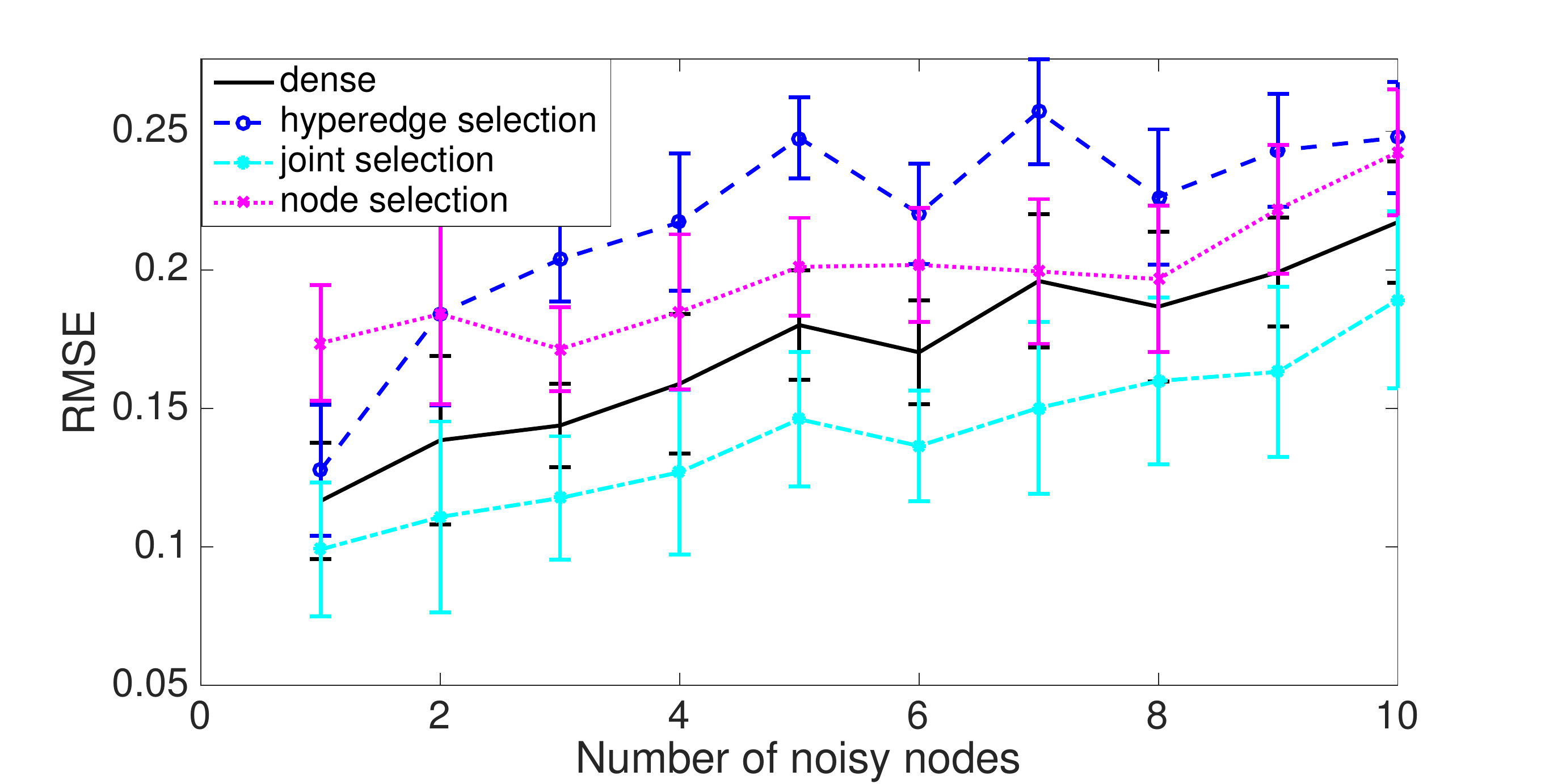}
	 \caption{Simulation data with both irrelevant hyperedges and noisy nodes. The data was generated with five irrelevant hyperedges. X axis was the different numbers of noisy nodes added to each relevant hyperedge and y axis was the RMSEs of the models. Joint selection gave the lowest RMSEs, showing its effectiveness for both irrelevant hyperedges and noisy nodes.}
  	\label{fig:sim2}
\end{figure}

We simulated this case by having five irrelevant hyperedges and different numbers of noisy nodes on each relevant hyperedge and carried out experiments in a similar manner. RMSE results were shown in Figure \ref{fig:sim2}. We found that joint hyperedge and node selection model  (\ref{starjointsel})  had the lowest  RMSEs consistently. It showed the ability of this model to select both hyperedges and nodes at the same time. All other models, dense, hyperedge selection and node selection models, were not as effective to handle this case. We would consider this was the most practical case as there would be irrelevant and noisy data in the dataset.
In this case, our proposed joint selection model was the most advantageous due to its flexibility to handle both irrelevant hyperedges and noisy nodes.

\subsection{Sparsity with $\lambda$}

\begin{figure}[t]
  \begin{center}
    \includegraphics[width=1\linewidth]{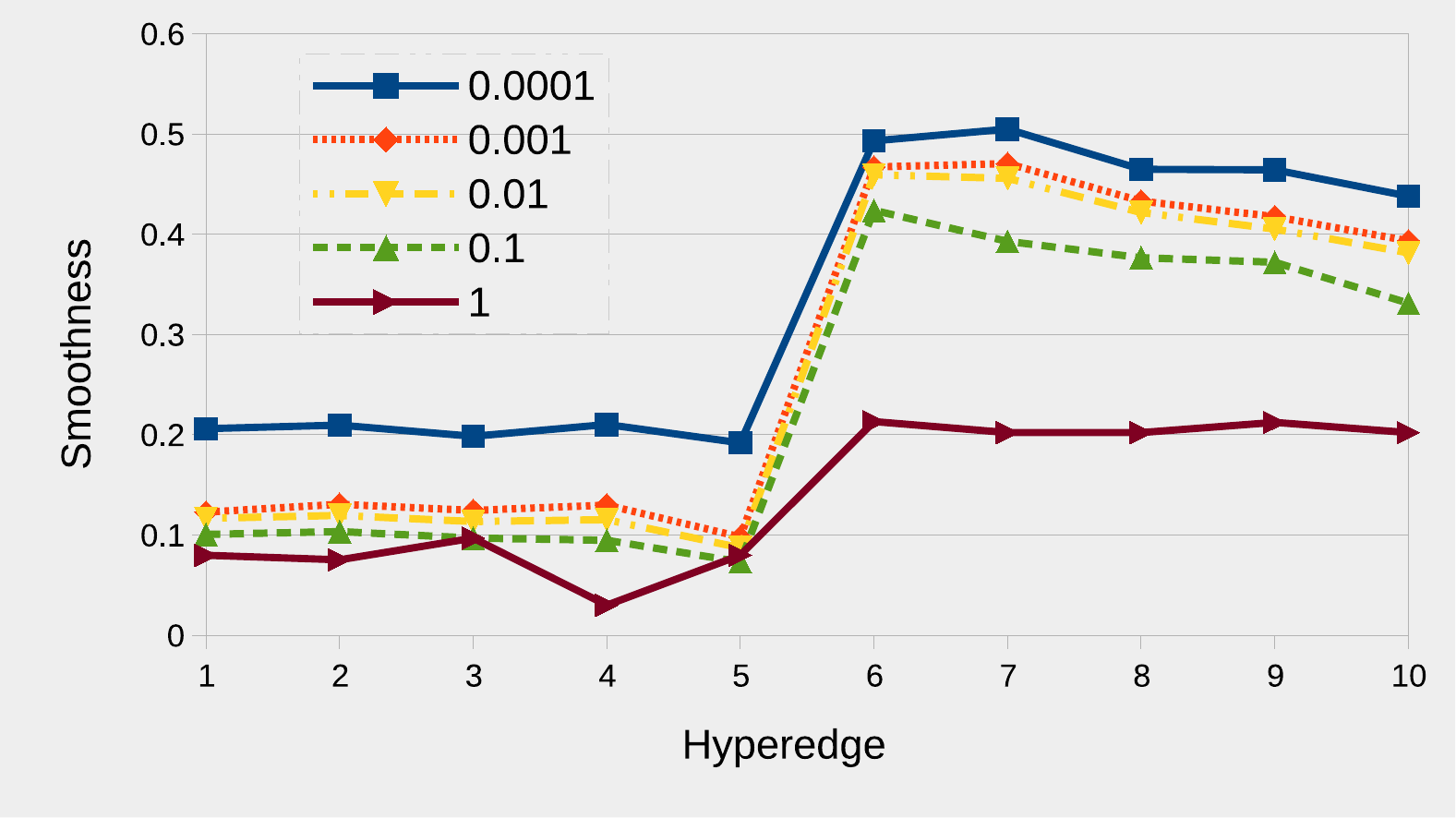}  
    \end{center}
 \caption{Simulation results on hypergraphs with five relevant hyperedges and five irrelevant hyperedges.  X axis was the indexes of hyperedges. From 1 to 5 were relevant hyperedges and from 6 to 10 were irrelevant ones. Y axis was the smoothness measures  on each hyperedge  with different $\lambda$. We could observe that when $\lambda$ were small enough, large gaps occured between smoothness measures of relevant and irrelevant hyperedge groups.} 
\label{fig:lambda}
\end{figure}

We also showed the sparsistency arguments  (\ref{properties}) on simulated data with five relevant/irrelevant edges with our proposed joint selection model. Data were split to train and test as above. We showed the smoothness measures ($\delta_k$) with different values of regularization parameter $\lambda$. It was proved in theory that the gap between smoothness measures of a relevant hyperedge and an irrelevant hyperedge is guaranteed if $\lambda$ was small enough. We could confirm the theory in Figure \ref{fig:lambda} that, as $\lambda$ decreased, smoothness measures on relevant hyperedges (indexed from 1 to 5) became significantly smaller than that of irrelevant hyperedges (indexed 6 to 10).

\section{Experiments}

\begin{table*}[t]
\caption{RMSEs of the four models on real data. $n$ and $m$ are the numbers of nodes and hyperedges, respectively. Bold numbers are the lowest RMSEs among the methods with 95\% statistical significance using t-test. The models with the lowest RMSEs are sparse models.
}
\label{tab:realdat}
\begin{center}
\begin{tabular}{|l|l|l|l|l|l|l|}
\hline
Dataset & $n$ & $m $ & Dense  & Hyperedge Select. & Node Select. & Joint Select.  \\ \hline
HayesRoth & 132 &15 & 0.587$\pm$0.044&0.600$\pm$0.071 & 0.758$\pm$0.076 & 0.746$\pm$0.067 \\
Lenses &	24 & 9 &  0.730$\pm$0.215 &	 \textbf{0.574$\pm$0.248}  & 0.767$\pm$0.227 & 0.770$\pm$0.227 \\
Congress 	& 435 & 48 &  0.373$\pm$0.011 &	 0.473$\pm$0.012  &0.444$\pm$0.010 &	 \textbf{0.306$\pm$0.034} \\
Spect &	267 & 44 &  0.384$\pm$0.035&0.400$\pm$0.021   &0.405$\pm$0.057 &	 0.404$\pm$0.031\\
TicTacToe & 958 & 27 & 	 0.468$\pm$0.009 &	 0.476$\pm$0.009  &  0.481$\pm$0.019 &	 0.476$\pm$0.009 \\ 
Car & 1728  & 21 & 	 0.692$\pm$0.043 &	 \textbf{0.462$\pm$0.026}  & 0.748$\pm$0.043 &	 0.740$\pm$0.044 \\
Monks &	124 & 17 &  0.469$\pm$0.008 &	 0.437$\pm$0.023 & 0.528$\pm$0.029 &	 0.504$\pm$0.004 \\
Balance &  625	 & 20 &   0.831$\pm$0.013&	 0.955$\pm$0.010 & 0.916$\pm$0.014 &	 \textbf{0.629$\pm$0.044} \\
 \hline
\end{tabular}
\end{center}
\end{table*}

We conducted experiments on real benchmark data to show the effectiveness of the four models in real situations. We also compared regression performances of these models on to show the benefits of learning sparsely smooth models on hypergraphs. We generated a hypergraph from multivariate categorical data as follows. For each categorical variable, each group of points having the same category formed a hyperedge. The set of all hyperedges (for all categories of all variables) formed the hypergraph. We chose multivariate categorical data that were publicly available in UCI machine learning repository \cite{Lichman13}. Parameter setting of this experiment was the same as in simulations.  We used ten-fold cross-validation to split train/test data and averaged RMSEs over ten runs. We collected the least RMSEs for different hyperparameter values from its predefined set (as in previous section).

Performances of the models were shown in Table \ref{tab:realdat}.  We observed that sparsely smooth models always have lower or equal RMSEs compared to the dense model.  In fact, four out of eight datasets showed the significantly lowest RMSEs, which are all from sparse models.This might be attributed to the fact that sparse models are usually robust to irrelevant hyperedges and noisy nodes. We could conclude that sparsely smooth models are general learning tools for hypergraphs and can be used to test for irrelevant hyperedges  or noisy nodes. 

\section{Conclusion}
The paper dealt with the problem of learning smooth functions on hypergraphs in the presence of irrelevant hyperedges and noisy nodes. We first proposed a general framework of smooth functions that includes all previous models. The framework could be used to derive many new models. To address the case of noisy and irrelevant data, we propose to incorporate sparse learning into hypergraph setting.
To deal with irrelevant hyperedges and noisy nodes, we proposed \emph{sparsely smooth formulations} to learn the functions that are smooth on a small number of hyperedges/nodes. We showed their sparse-inducing properties and analyzed their sparse support recovery (sparsistency) results,  We found that our proposed joint selection model was able to recover the correct sparse support while hyperedge selection model was not. We then conducted experiments on simulated and real data to validate their properties and performances. Sparsely smooth models  were  shown to be more robust in the presence of irrelevant hyperedges and noisy nodes. In summary, our proposed models guaranteed both theoretical soundness and practical effectiveness. We concluded that we have added sparsity into learning smooth models on hypergraphs.

\bibliographystyle{IEEEtran}
\bibliography{IEEEabrv,hyperLong}

\begin{IEEEbiography}{Canh Hao Nguyen}
Canh Hao Nguyen received his B.S. degree in Computer Science from the University of New South Wales, Australia, M.S. and Ph.D. degrees from JAIST, Japan. He has been working in machine learning and Bioinformatics. His current interests are machine learning for graph data with applications in biological networks. 
\end{IEEEbiography}

\begin{IEEEbiography}{Hiroshi Mamitsuka}
Hiroshi Mamitsuka received the B.S. degree in biophysics and
biochemistry, the M.E. degree in information engineering, and the
Ph.D. degree in information sciences from the University of Tokyo,
Tokyo, Japan, in 1988, 1991, and 1999, respectively. He has been
working on research in machine learning, data mining, and
bioinformatics. His current research interests include mining from
graphs and networks in biology and chemistry.
\end{IEEEbiography}

\end{document}